%% file: aistats2018.tex
\documentclass[twoside]{article}

\newif\ifisarxiv
\isarxivtrue
\newif\ifcitesupp
\citesuppfalse

\ifisarxiv
\usepackage[accepted]{aistats2018_arxiv}
\else
\usepackage[accepted]{aistats2018}
\fi

\usepackage[utf8]{inputenc} 
\usepackage[T1]{fontenc}    
\usepackage{hyperref}       
\usepackage{url}            
\usepackage{booktabs}       
\usepackage{amsfonts}       
\usepackage{nicefrac}       
\usepackage{microtype}      

\usepackage{times}
\usepackage{graphicx} 
\usepackage{subfigure} 
\usepackage{color}

\usepackage{amsmath}
\usepackage{verbatim}
\usepackage{wrapfig}

\usepackage{algorithm}
\usepackage{algorithmic}
\usepackage{enumitem}

\usepackage{forloop}
\newcounter{loopcntr}

\definecolor{darkSilver}{cmyk}{0,0,0,0.1}

\ifisarxiv
\input{defs}

\else
\input{../defs}
\fi
\newtheorem{theoreme}{Theorem} 
\newtheorem{proposition}[theoreme]{Proposition}
\newtheorem{lemma}[theoreme]{Lemma}
\newtheorem{definition}[theoreme]{Definition}
\newtheorem{corollary}[theoreme]{Corollary}
\newtheorem{remark}[theoreme]{Remark}
\newtheorem{example}[theoreme]{Example}

\newtheorem{conjecture}{Conjecture}

\begin{document}

\setlength{\abovedisplayskip}{6pt}
\setlength{\belowdisplayskip}{6pt}
%
\runningtitle{Batch-Expansion Training:
  An Efficient Optimization Framework}

%
\runningauthor{Derezi\'{n}ski, Mahajan, Keerthi, Vishwanathan, Weimer}

\twocolumn[

\aistatstitle{Batch-Expansion Training:\\
  An Efficient Optimization Framework$^*$}

\aistatsauthor{Micha{\polishl } Derezi\'{n}ski\\
Dept. of Computer Science\\
UC Santa Cruz\\
\texttt{mderezin@ucsc.edu}
\And Dhruv Mahajan\\
Facebook Research\\
Menlo Park\\
\texttt{dhruvm@fb.com}
 \And S. Sathiya Keerthi\\
Microsoft Corporation\\
Mountain View\\
\texttt{keerthi@microsoft.com}
\AND S. V. N. Vishwanathan
\And Markus Weimer}


\aistatsaddress{
Dept. of Computer Science\\
UC Santa Cruz\\
\texttt{vishy@ucsc.edu}
\And 
Microsoft Corporation\\
Mountain View\\
\texttt{mweimer@microsoft.com}
} ]

\begin{abstract}
  We propose Batch-Expansion Training (BET), a framework for running a
  batch optimizer on a gradually expanding dataset. As opposed to
  stochastic approaches, batches do not need to be resampled i.i.d. at every
  iteration, thus making BET more resource efficient in a distributed
  setting, and when disk-access is constrained. Moreover, BET can be
  easily paired with most batch optimizers, does not require any
  parameter-tuning, and compares favorably to existing stochastic and
  batch methods. We show that when the batch size grows exponentially
  with the number of outer iterations, BET achieves optimal
  $\Ocalt(1/\epsilon)$ data-access convergence rate for strongly convex
  objectives. Experiments in parallel and distributed settings show
  that BET performs better than standard batch and stochastic
  approaches. 
\end{abstract}


{\renewcommand{\thefootnote}{\fnsymbol{footnote}}
\footnotetext[1]{This research was supported by NSF grant IIS-1546452.}}

\section{Introduction}
\label{sec:introduction}
\input{introduction}
\section{Related Work}
\label{sec:related-work}
\input{related-work}

\section{Batch-Expansion Training}
\label{sec:expanding-batch-training}
\input{expanding-batch-training}


\section{Complexity Analysis}
\label{sec:complexity-analysis}
\input{complexity-analysis}

\section{Experiments}
\label{sec:experiments}
\input{experiments}

\section{Conclusions}
\label{sec:conclusions}
\input{conclusions}


\bibliographystyle{plain}
\bibliography{aistats2018}


\clearpage
\newpage

\appendix

\section{Proof Details for Theorem \ref{thm:complexity}}
\label{sec:thm-proof}

\input{lemma-proof}

\section{Additional Experiments}
\label{sec:additional-experiments}

\input{additional-experiments}

\end{document}

%% file: defs.tex
\DeclareMathOperator{\x}{\mathbf{x}}

\edef\polishl{\l}

\RequirePackage{latexsym}
\RequirePackage{amsmath}
\RequirePackage{amssymb} 
\RequirePackage{color} 
\RequirePackage{bm}
\RequirePackage{color}
\RequirePackage{picinpar}

\newcommand{\fh}{\hat{f}}

\newcommand{\gh}{\hat{g}}

\DeclareMathOperator{\wb}{\mathbf{w}}
\DeclareMathOperator{\wbh}{\mathbf{\widehat{w}}}
\DeclareMathOperator{\xb}{\mathbf{x}}

\renewcommand{\l}{\ell}

\DeclareMathOperator{\Wb}{\mathbf{W}}

\DeclareMathOperator{\Zb}{\mathbf{Z}}

\DeclareMathOperator{\Hcal}{\mathcal{H}}

\DeclareMathOperator{\Ocal}{\mathcal{O}}
\DeclareMathOperator{\Ocalt}{\tilde{\mathcal{O}}}

\DeclareMathOperator{\Qcal}{\mathcal{Q}}

\DeclareMathOperator{\Rcal}{\mathcal{R}}

\ifx\BlackBox\undefined
\newcommand{\BlackBox}{\rule{1.5ex}{1.5ex}}  
\fi

\ifx\proof\undefined
\newenvironment{proof}{\par\noindent{\bf Proof\ }}{\hfill\BlackBox\\[2mm]}
\fi

\newtheorem{theorem}{Theorem}[section]

\ifx\Brown\undefined
\definecolor{brown}{rgb}{0.5,0.1,0.1}
\newcommand{\Brown}[1]{\color{brown}{#1}\color{black}}
\fi

\ifx\Red\undefined
\definecolor{red}{rgb}{1.0,0,0}
\newcommand{\Red}[1]{\color{red}{#1}\color{black}}
\fi

\ifx\Green\undefined
\definecolor{green}{rgb}{0,0.4,0}
\newcommand{\Green}[1]{\color{green}{#1}\color{black}}
\fi

\ifx\Blue\undefined
\definecolor{blue}{rgb}{0,0,1.0}
\newcommand{\Blue}[1]{\color{blue}{#1}\color{black}}
\fi

\newcommand{\beq}{\begin{equation}}
\newcommand{\eeq}{\end{equation}}
\newcommand{\beh}{\begin{conjecture}}
\newcommand{\eeh}{\end{conjecture}}
\newcommand\bel{\begin{lemma}}
\newcommand\eel{\end{lemma}}
\newcommand\bet{\begin{theorem}}
\newcommand\eet{\end{theorem}}
\newcommand\bex{\begin{example}}
\newcommand\eex{\end{example}}
\newcommand\bed{\begin{definition}}
\newcommand\eed{\end{definition}}
\newcommand\bep{\begin{proposition}}
\newcommand\eep{\end{proposition}}
\newcommand\ber{\begin{remark}}
\newcommand\eer{\end{remark}}
\newcommand\bec{\begin{corollary}}
\newcommand\eec{\end{corollary}}
\newcommand\qed{\hfill$\Box$\medskip}
\newcommand\cH{{\mathcal H}}

\def\zz{{\mathbb Z}}

\def\ee{{\mathbb E}}


%% file: introduction.tex
State-of-the-art optimization algorithms used in machine learning
broadly tend to fall into two main categories: batch methods, which
visit the entire dataset once before performing an expensive parameter
update, and stochastic methods, which rely on a small subset of training
data, to apply quick parameter updates at a much greater frequency. Both
approaches present different trade-offs. Stochastic updates often
provide very good early performance, since they can update the
parameters a considerable number of times before even a single batch
update finishes. On the other hand, by accessing the full dataset at
each step, batch algorithms can better utilize second-order information
about the loss function, while taking advantage of parallel and
distributed architectures. Finding approaches that provide the
best of both worlds is an important area of research. In this paper, we
propose a new framework which\,--\,while closer to batch in
spirit\,--\,enjoys the benefits of stochastic methods. In addition, our
framework addresses a practical issue observed in real-world industrial
settings, which we now describe.

In industrial server farms, compute resources for large jobs become
available only in a phased manner. When running a batch optimizer, which
requires loading the entire dataset, one has to wait until all the
machines become available before beginning the computation. Moreover,
the training data, which typically consists of user logs, is distributed
across multiple locations. This data needs to be normalized, often by
communicating summary statistics or subsets of the data across the
network. Since stochastic optimization algorithms only deal with a
subset of the data at a time, one can largely avoid the bottleneck of
data normalization by preparing mini-batches at a time. However, now
each data point needs to be visited multiple times randomly, and this
requires performing many random accesses from a hard-disk or network
attached storage (NAS), which is inherently slow. Moreover, extending
stochastic optimization to the distributed setting is still an active
area of research. This raises the question of whether the compute and
data-availability delays could be avoided in batch methods, 
and how that would affect the training time.


We propose Batch-Expansion Training (BET), an adaptive, parameter-free 
meta-algorithm, which can be paired with most batch optimization
methods to accelerate their performance in sequential, parallel, as
well as distributed settings, addressing the issues discussed
above. Our approach hinges on the following 
observation: initially, when only a subset of the data is available, the
statistical error (the error that arises because one is observing a
sample from the true underlying data distribution) is large. Therefore,
we can tolerate a large optimization error (the error that arises
because of the iterative nature of the batch optimizer). However, as the
sample size increases, the statistical error decreases and the
corresponding optimization error that we can tolerate also
decreases. BET exploits this by initially training models with large
optimization error on smaller subsets of the data, and iteratively
loading more training data, and driving down the optimization error.

Relying on classical results in statistical learning theory, we show
that any optimizer exhibiting linear convergence rate can be
effectively accelerated by periodically doubling the data size
after a certain number of iterations. We propose a simple
parameter-free algorithm, which dynamically decides the optimal number
of iterations between each data expansion, adapting to the performance
of a batch optimizer provided as a black box method. Experiments using
Nonlinear Conjugate Gradient (CG), as well as the L-BFGS method,
demonstrate the versatility of our framework. We show that for strongly
convex losses BET achieves the same asymptotic $\Ocalt(1/\epsilon)$
convergence rate as SGD in terms of data accesses (and is strictly
faster than regular batch updates). However, unlike stochastic methods,
BET reuses all of the data that has already been loaded, which helps
it scale very well in parallel and distributed settings, as confirmed
by the experimental results.

%% file: related-work.tex
There has been a recent explosion of interest in both batch and
stochastic optimization methods for machine learning.
Algorithms like Stochastic Variance Reduced Gradient method
\cite{svrg} and related approaches \cite{sag,saga} mix SGD-like steps
with some batch computations to control the stochastic noise. Others
have proposed to parallelize stochastic training through large
mini-batches \cite{efficient-mini-batch,distributed-mini-batch}.
However, these methods do not address the issues of compute and
data-availability delays that we discussed above. 
Additionally, two-stage approaches have been proposed
\cite{terascale,sdca-convergence}, which employ SGD at the beginning
followed by a batch optimizer (e.g., L-BFGS). These methods are much
more limited than BET, which allows for multiple stages of optimization,
with clear practical benefits.
Interleaving computation with data loading was shown to have significant
practical benefits by \cite{stream-svm}. However, that work is confined
to training on a single machine and did not provide any theoretical
convergence guarantees. In contrast, we largely focus on the distributed
setting and provide convergence guarantees.

Several works explore the idea of batch expansion in various
forms. \cite{dynamic-samples,hybrid-deterministic-stochastic} propose
using stochastic mini-batches of gradually increasing size paired with
optimizers like gradient descent or Newton-CG. The algorithms proposed there
strongly rely on stochastic sampling, which makes them less
resource-efficient than BET, and not applicable to certain distributed
settings. The convergence guarantees offered in
\cite{dynamic-samples} are limited to using gradient descent as the
inner optimizer, and heavily rely on the independence conditions
present in stochastic sampling. 
The idea of gradually increasing batch size without resampling the
batches was first discussed by \cite{warm-start}, however their
convergence analysis is also limited to gradient descent.
Recently, \cite{adaptive-newton,starting-small}
proposed variants of the Newton's method and of the SAGA \cite{saga}
algorithm which run on increasing batch sizes. 
Both of those methods pose challenges in large-scale distributed
settings (expensive hessian computation required for Newton's
method and sequential stochastic updates in SAGA). The key advantage of our
meta-algorithm is that it can be used with any inner batch optimizer
as a black box, including quasi-Newton methods like L-BFGS, making
this approach more broadly applicable and scalable. Similarly,
our complexity analysis 
applies to a wide range of inner optimizers.


%% file: expanding-batch-training.tex
We consider a standard composite convex optimization problem arising
in regularized linear prediction. Given a dataset $\Zb =
\{(\x_i,y_i)\}_{i=1}^N$, we aim to approximately minimize the average
regularized loss
\vspace{-1mm}
\begin{align}
\fh \triangleq \frac{1}{N}\sum_{i=1}^N
\ell_{z_i}(\wb) + \frac{\lambda}{2}\|\wb\|^2,\label{eq:loss}
\end{align}
\vspace{-4mm}

where $z_i = (\x_i,y_i)$ and
$\ell_{z_i}(\wb)\triangleq\ell(\langle\wb,\phi(\x_i)\rangle,y_i)$ is
the loss of predicting with 
a linear model $\langle\wb,\phi(\x_i)\rangle$ against a target label
$y_i$. Any iterative optimization algorithm in this setting will
produce a sequence of models $\{\wb_t\}_{t=1}^T$ with the goal that
$\wb_T$  has small optimization error $\gh(\wb_T)$ with
respect to the exact optimum $\wbh^*$, where 
\[\gh(\wb)\triangleq \fh(\wb) - \fh(\wbh^*)\quad \textnormal{and}
\quad\wbh^* \triangleq \arg\min_{\wb}\ \fh(\wb).\] 
Note that our true goal is to predict well on
an unseen example $z =(\x,y)$ coming from an underlying
distribution. 
The right regularization $\lambda$ for this task can be determined
experimentally or 
from the statistical guarantees of loss function $\ell$, as discussed in
\cite{fast-rates,svm-datasize}. In this paper, however, we will assume
that an acceptable $\lambda$ was chosen and concentrate on the problem
of minimizing $\fh$. 

\subsection{Linear convergence of the batch optimizer}
\label{sec:linear-optimizer}
Adding $\ell_2$-norm regularization in function $\fh$ makes it
$\lambda$-strongly convex.
In this setting, many popular batch optimization algorithms
enjoy linear convergence rate \cite{nocedal-wright-book}. Namely, given arbitrary model $\wb$ and
any $c>1$, after at
most $\Ocal(\log(c))$ iterations - where a single iteration can look
at the entire dataset -  we can 
reduce its optimization error by a multiplicative factor of $c$, obtaining
$\wb'$ such that
\[\gh(\wb')\leq c^{-1}\cdot \gh(\wb).\]
Note that the runtime of a single iteration will depend on data size
$N$, but the number of needed iterations does not. 
Setting $c=2$, we
observe that only a constant number of batch iterations is necessary
for a linearly converging  
optimizer to halve the optimization error of $\wb$. This constant
depends on the convergence rate enjoyed by the method. For
example, in the case of batch gradient descent we need $\Ocal(1/\lambda)$
iterations to reduce the optimization error by a factor of 2
\cite{bubeck15} (note the dependence on the strong convexity 
coefficient), while other methods (like L-BFGS) can achieve better
rates of convergence \cite{trust-region-large-scale}. Furthermore,
the time complexity of performing a single 
iteration for many of those algorithms (including GD and L-BFGS) is
linearly proportional to the data size. We refer to 
methods exhibiting both linear convergence (with
respect to a given loss) and linear time complexity of a single
iteration as {\em linear optimizers}. From now on, we only consider 
this class of methods.

\subsection{Dataset size selection}

\begin{figure}
\begin{center}
\includegraphics[width=0.5\textwidth]{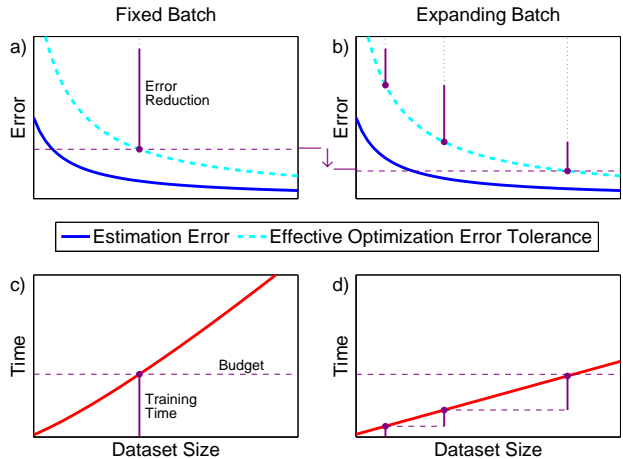}
\end{center}
\caption{Estimation error and effective optimization error tolerance are inversely
  proportional to the dataset size (plots a and b). Fixed Batch
  preselects a data size, then reduces the optimization error
  (vertical line in a), whereas Expanding 
  Batch divides the work into stages with different data sizes
  (vertical lines in b). Optimization is faster for smaller data
  sizes (see plots c and d), so Expanding Batch reaches smaller
  effective optimization error tolerance within the same time budget
  compared to Fixed Batch (plots a and b).
}
\label{fig:data-doubling}
\end{figure}
The general task of an optimization algorithm is to return a model
with small optimization error $\gh(\wb)\leq \epsilon$.
For selecting an effective optimization error
tolerance $\epsilon$ in a machine learning problem,
we often look at the
estimation error exhibited by the objective, i.e. how much $\fh$
deviates from the expected regularized loss measured on a random unseen
example. As discussed in \cite{svm-datasize,large-scale-tradeoffs}, the effective
optimization error tolerance should be proportional to the 
 estimation error of $\fh$, because optimizing beyond that point does
 not yield any improvement on unseen data. Note that under standard
 statistical assumptions, the more data we use, the 
 smaller the estimation error becomes \cite{fast-rates}, and thus,
 the effective optimization error tolerance should also be decreased
 (see Figure \ref{fig:data-doubling}a).

Consider the scenario where data is abundant, but we have a limited
time budget for training the model\footnote{This is a reasonable
practice in many web applications.}. In this
case, a practitioner would select the data size so that we can
reach the smallest effective optimization error tolerance (dashed curve
in Figure \ref{fig:data-doubling}a) within the allotted training
time. Vertical line in Figure \ref{fig:data-doubling}a shows the
reduction in optimization error that the algorithm has to achieve to
obtain the desired tolerance.
 If we use a batch optimizer for this task, the training time is
 determined by two factors. First, as the dataset grows, each
iteration takes longer (e.g. for linear optimizers the iteration time is
proportional to the data size, as discussed in Section
\ref{sec:linear-optimizer}). Second, the 
algorithm has to perform larger number of iterations, the smaller the
effective optimization error tolerance. Combined, those two effects result in
training time, which, for linear optimizers, grows faster than
linearly with the data size (see Figure \ref{fig:data-doubling}c).
 

In this paper, we propose that instead of finding the optimal dataset
size at the beginning, we first load a small subset of data,
train the model until we reach the
corresponding effective optimization error tolerance, then we load more data, and optimize
further, etc. Vertical lines in Figure \ref{fig:data-doubling}b
illustrate how optimization error is reduced in multiple
stages working with increasing data
sizes. This procedure benefits from faster iterations in the early
stages, as shown in Figure \ref{fig:data-doubling}d, with vertical
lines corresponding to the training time for each data size, and the
horizontal line showing the time budget (note that the
budget shown for this method is the same as the one used in
Figure \ref{fig:data-doubling}c). Thus, our
approach is able to reach smaller effective optimization error tolerance
within the same time budget compared to fixing the dataset size at the
beginning (see Figures \ref{fig:data-doubling}a and \ref{fig:data-doubling}b).

\subsection{Exponentially increasing batches}
\label{sec:exponentially-increasing-batches}

We now precisely formulate the idea of training with gradually
increasing data size. Suppose our goal is to return a model
$\wbh$ with optimization error $\gh(\wbh)\leq\epsilon$.
In the procedure described above, we seek to obtain a sequence of
gradually improving models $\wb_1,\ldots, \wb_{T-1}$ before we reach 
$\wb_T=\wbh$. Any of the intermediate models (say, model
$\wb_t$ for $t<T$) has a large optimization error relative to
$\wbh$, so to compute $\wb_t$ we can minimize
an objective function with correspondingly large estimation error.
Thus, we will first obtain $\wb_1$ with
optimization error\footnote{The optimization error for intermediate models
$\wb_t$ is computed using only the loaded portion of the data.}
$\epsilon_1$ using $n_1$ data points, next we pick 
a smaller error tolerance $\epsilon_2<\epsilon_1$ and bigger data size $n_2>n_1$,
computing a better model $\wb_2$, etc. so that in the end we reach
$\epsilon_T=\epsilon$. Algorithm \ref{alg:simple} demonstrates
a simple instantiation of this strategy, where at each stage we double
the data size:
\begin{algorithm}[H]
  \caption{Batch-Expansion Training}
  \begin{algorithmic} [1]
    \STATE Pick initial model $\wb_0$
    \STATE Load first $n_1$ data points
    \FOR {$t=1..T$}
      \STATE $\wb_{t}\leftarrow$ run $\kappa_t$
      iterations on $n_t$ data points \\
      \STATE $n_{t+1}\leftarrow b_tn_t$ \hfill (increase data size)
      \STATE Load $n_{t+1}-n_t$ data points
    \ENDFOR
    \RETURN $\wb_T$
 \end{algorithmic}
\label{alg:simple}
\end{algorithm}
\vspace{-3mm}
The basis for the number of inner iterations $\kappa_t$
will be explained below.
Note that at stage $t$ of the process we are working with an estimate
of the loss function  
\[\fh_t(\wb)\triangleq \frac{1}{n_t}\sum_{i=1}^{n_t}\ell_{z_i}(\wb) +
\frac{\lambda}{2}\|\wb\|^2,\]
which tends to $\fh$ with increasing $t$. At this stage, the optimizer is
in fact converging to an approximate optimum
\[\wbh^*_t\triangleq \arg\min_w\fh_t(\wb),\]
rather than to the minimizer of $\fh$, $\wbh^*$. 
To decide how much data is needed at each stage we will describe the
relationship between data size $n_t$ and the desired optimization error
$\epsilon_t$. Note that model $\wb_t$ obtained at stage $t$ is
assumed to satisfy $\gh_t(\wb_t)\leq \epsilon_t$, where $\gh_t$
represents optimization error for the given data subset, i.e. 
$\gh_t(\wb) = \fh_t(\wb) -\fh_t(\wbh_t^*).$ When going to the next
stage, the data size increases, and thus we can use
optimization error function $\gh_{t+1}$, which is a
better estimate of $\gh$ than $\gh_t$ is.
In Appendix \ref{sec:lemma-proof}%
\ifcitesupp~ of \cite{supplement}\fi,
we show that function $\gh_{t+1}$ can be uniformly bounded by
$\gh_t$, plus an
additional term which can be interpreted as the estimation error
(of $\gh_t$ with respect to $\gh_{t+1}$):
\begin{align}
\gh_{t+1}(\wb)\leq 2\cdot \gh_t(\wb) + \Ocal\left( 1/(\lambda
    n_t)\right).\label{eq:lemma-repeat}
\end{align}
As discussed earlier, the effective optimization error for data size
$n_t$ is proportional to the estimation error suffered by $\gh_t$,
hence it is sufficient to demand that 
\begin{align}
 \gh_t(\wb_t)\leq \epsilon_t\triangleq
 \Ocal\left(1/(\lambda n_t)\right).\label{eq:epsilont-nt}
\end{align}
Note that this makes the optimization error tolerance $\epsilon_t$ inversely proportional
to the subset size $n_t$, confirming our intuition that for $t<T$, 
since $\epsilon_t>\epsilon$, we can work with a batch of size $n_t$
that is smaller than $N$. 


To establish the correct rate of growth for the data size
$n_t$, we combine (\ref{eq:epsilont-nt}) with the
observations made in Section \ref{sec:linear-optimizer}.
First, note 
that the rate of decay of sequence $\{\epsilon_t\}$ should match 
the convergence rate of the optimization algorithm, so that inequality
$\gh_t(\wb_t)\leq\epsilon_t$ can be satisfied for all $t$.
Recall from Section \ref{sec:linear-optimizer}, that a
linear optimizer takes only a constant number of
iterations to reduce the optimization error by a
constant factor. Suppose that improvement by a factor of $6$ takes
$\kappa$ iterations (going from $\wb_t$ to $\wb_{t+1}$). Then, using
(\ref{eq:lemma-repeat}) and (\ref{eq:epsilont-nt}) we have
\begin{align*}
\gh_{t+1}(\wb_{t+1}) &\leq \frac{\gh_{t+1}(\wb_t)}{6} \leq
\frac{2\cdot \gh_t(\wb_t) + \epsilon_t}{6}\leq 
\frac{\epsilon_t}{2}.
\end{align*}
This suggests the following simple strategy: 
at each stage perform $\kappa$ iterations of the
optimizer, then divide the tolerance level by 2 (matching the
convergence rate). Note that based on 
Equation (\ref{eq:epsilont-nt}), this corresponds to doubling the data
size $n_t$. Thus, we obtain the following simple scheme for data
expansion, which maintains the desired relationship between $n_t$ and
$\epsilon_t$:
\[\epsilon_{t+1} = \epsilon_t/2,\qquad n_{t+1} = 2\cdot n_t. \]
It is important that $n_t$ grows exponentially with $t$.
This allows for a considerable
improvement in runtime. Let us say that $\epsilon_0=\Ocal(1)$ and $T$
stages are needed to reach the final desired tolerance $\epsilon$,
i.e. $T=\log(\epsilon_0/\epsilon) =\Ocal(\log(1/\epsilon))$. Moreover,
using (\ref{eq:epsilont-nt}) the suitable size of the full dataset is
$N=\Ocal(1/(\lambda\epsilon))$. If we assume that one iteration of  
the linear optimizer takes time proportional to the data size, then
the time complexity of the optimization when using batch-expansion is
given by 
\[\sum_{t=1}^T \kappa n_t = \kappa n_0\sum_{t=1}^T2^t = \Ocal(\kappa
N) = \Ocal\left(\frac{\kappa}{\lambda\epsilon}\right). \]
On the other hand, when running the same optimizer on full dataset from the
beginning, the time complexity becomes
\[ \Ocal(\kappa N\cdot T) =
\Ocal\left(\frac{\kappa}{\lambda\epsilon}\cdot\log(1/\epsilon)\right).\] 
Note that to establish convergence of the proposed algorithm,
it is only required that the dataset is randomly permuted, i.e. that each
subset $\{z_i\}_{i=1}^{n_t}$ represents a random portion of the
data. However, the batches used in different stages do not need to be
independent of each other, which is why we can reuse data from
previous stages. Section \ref{sec:complexity-analysis}
precisely formulates the ideas discussed above, and a careful
analysis of the time complexity is given in Theorem
\ref{thm:complexity}. 

\subsection{Two-track algorithm}
How many iterations of the proposed expansion procedure should be
performed at each stage? From our high-level analysis, we concluded 
that a roughly constant number of updates should be sufficient for
any stage (using a linear optimizer), since each time we aim to make the
same multiplicative improvement to the optimization error suffered by our
model. However, that constant may depend on the type of loss function,
the dataset, as well as the optimizer used, and moreover, in practice the
right number of iterations may in fact vary to some extent between the
stages. Therefore, we need a practical method of deciding the right
time to double the data.
 Consider the following experiment: we
run two optimization tracks in parallel, first one for the batch of
size $n_t$, the other for half of that batch. 
One update on the bigger batch takes longer than an update on the
smaller one. Which
track will make better progress towards the optimum of the bigger
batch in the same amount of time? If the starting model is far enough
from the optimum $\wbh^*_t$, then the faster updates will initially
have an advantage. However, as the 
convergence proceeds, only the slower track can get arbitrarily close,
so at some point it will move ahead of the fast one (in terms of
the loss $\fh_{t}(\wb)$). Denote the starting model as
$\wb_{t,0}$. The secondary track (running on half of the batch) also
starts at the same point, denoted as $\wb'_{t-1,0}=\wb_{t,0}$, and
they are both updated:
\begin{align*}
\wb_{t,s+1} &\leftarrow \textrm{Update}(\wb_{t,s},n_t),\\
\wb'_{t-1,s+1} &\leftarrow \textrm{Update}(\wb'_{t-1,s},n_{t-1}),
\end{align*}
where $\textrm{Update}(\wb,n)$ is one step of the optimizer, with
respect to model $\wb$, on the batch $\{z_i\}_{i=1}^n$. 
Let $c_s$ and $c'_s$ be the total runtimes of the first and second track
after $s$ iterations, respectively, and let $s_1(s)=\max\{s_0 :
c_{s_0}\leq c'_s\}$ be the number of iterations that the slower
track completes in the time it took the faster track to do $s$.
To compare the performance of the two tracks we will use the
condition:
\begin{align}
\fh_t(\wb_{t,s_1(s)}) < \fh_t(\wb'_{t-1,s}).\label{eq:condition}
\end{align}
\vspace{-6mm}
\begin{algorithm}[H]
  \caption{Two-Track Optimizer}
  \begin{algorithmic} 
    \STATE Initialize $\wb_{1,0}=\wb'_{0,0}$ arbitrarily, $s\leftarrow
    0$, $t\leftarrow 1$
    \STATE Pick any $2\leq n_1=2 n_0< N$
    \WHILE {$n_t < N$}
    \STATE 
    \begin{tabular}{ll}
      $\wb_{t,s+1}$ & $ \leftarrow \textrm{Update}(\wb_{t,s},n_t)$\\
      $\wb'_{t-1,s+1}$ & $ \leftarrow    \textrm{Update}(\wb'_{t-1,s},n_{t-1})$\\  
      $s$ & $\leftarrow s+1$\\
      $s_1$& $\leftarrow\max\{s_0 : c_{s_0}\leq c'_s\}$
    \end{tabular}
    \IF {$\fh_t(\wb_{t,s_1}) <
      \fh_t(\wb'_{t-1,s})$}
        \STATE  
        \begin{tabular}{ll}
          $n_{t+1}$ & $\leftarrow 2 n_t$\\
          $\wb'_{t,0}, \wb_{t+1,0}$ & $ \leftarrow \wb_{t,s}$\\
          $t$ & $\leftarrow t+1$\\
          $s$ & $\leftarrow 0$
        \end{tabular}
    \ENDIF
    \ENDWHILE
    \WHILE {stopping condition not met}
      \STATE 
        \begin{tabular}{ll}
          $\wb_{t,s+1}$ & $ \leftarrow
          \textrm{Update}(\wb_{t,s},N)$\\
          $s$ & $\leftarrow s+1$
        \end{tabular}          
     \ENDWHILE
     \RETURN $\wb_{t,s}$
  \end{algorithmic}
  \label{alg:two-track}
\end{algorithm}
\vspace{-4mm}
Algorithm \ref{alg:two-track} describes
Batch-Expansion Training implemented using the Two-Track
strategy. Note that in this algorithm we run the two tracks
sequentially, in an alternating order, however running them
in parallel would further improve the overall performance.

\subsection{Discussion}
We found that the choice to increase data size by a factor of $2$ at
each stage (rather than by a different factor) is
not crucial for the optimization performance
(both theoretically and in practice),  therefore this
parameter does not require tuning. The initial subset size $n_0$ also
does not affect
performance significantly - generally, the larger $n_0$ we
select, the more updates will be performed before first
data expansion, but as long as the initial subset is
small enough, total optimization time will remain close to optimal.
Thus, Algorithm \ref{alg:two-track} does not require any
tuning to achieve good performance. Moreover, our method
 can be paired with many popular batch
optimizers, and it will automatically adapt its behavior to
the selected inner optimizer, as shown in Section 
\ref{sec:inner-optimizers}. 

It is important to note that the fraction of data accessed by the
algorithm is only gradually expanded as optimization proceeds. Moreover, BET
iterates multiple times over the data points that have 
already been loaded. Thus, it is very resource efficient in a way that
can be beneficial with:

\textbf{Slow disk-access}. Loading data from disk
  to memory can be a significant bottleneck \cite{stream-svm}.
  Performing multiple
  iterations over the data points while extra data is being loaded
 in parallel  provides speed-up.

\textbf{Resource ramp-up}. In distributed computing, often not all
  resources are made available immediately at the beginning of the
  optimization \cite{resource-elastic}, which similarly leads to gradual
  data availability. 



%% file: complexity-analysis.tex
In this section, we provide theoretical guarantees for the time
complexity of Batch-Expansion Training and compare them to other
approaches. 
For the remainder of this section, we assume that the inner optimizer
for some $\kappa>1$ and for every $t,\wb$ exhibits linear
convergence:
\begin{align*}
\gh_t(\text{Update}(\wb,n_t))\leq
\left(1-(1/\kappa)\right)\gh_t(\wb).
\end{align*}
\vspace{-7mm}


For the sake of complexity analysis, we discuss a parameterized
variant of our approach, described in
Algorithm \ref{alg:optimal}, and establish complexity results for it.
Here, the number of
updates needed at each stage is a fixed parameter $\hat{\kappa}$.


\begin{algorithm}
  \caption{Optimal BET}
  \begin{algorithmic} 
    \STATE {\bf Input}: Target tolerance $\epsilon$
    \STATE Pick $\epsilon_0$, $n_0$
    \STATE Initialize $\wb_0\leftarrow 0$, \ \ $t\leftarrow 0$
    and $\hat{\kappa} = \lceil \kappa \log(6)\rceil$
    \WHILE {$3\cdot \epsilon_t > \epsilon$}
      \STATE $n_{t+1} \leftarrow 2n_t$
\STATE $\wb_{t,0}\leftarrow \wb_t$
\FOR {$s=1..\hat{\kappa}$}
\STATE $\wb_{t,s}\leftarrow \text{Update}(\wb_{t,s-1},n_{t+1})$
\ENDFOR
\STATE $\wb_{t+1}\leftarrow \wb_{t,\hat{\kappa}}$
      \STATE $\epsilon_{t+1} \leftarrow \epsilon_t/2$
      \STATE $t \leftarrow t+1$
    \ENDWHILE
    \STATE $T\leftarrow t$
    \RETURN $\wb_T$
\end{algorithmic}
\label{alg:optimal}
\end{algorithm}

{\bf Assumptions.} We assume that the feature mapping $\phi(\cdot)$ (see
(\ref{eq:loss}) and the discussion below it) is $B$-bounded,
i.e. it satisfies
$\|\phi(\xb_i)\|\leq B$, and that the loss function
$\ell_{z_i}(\wb)=\ell(z_i,\wb)$ is $L$-Lipschitz in $z_i$ for all
$\wb$. Moreover, we will use the fact that $\fh$ is $\lambda$-strongly
convex, due to the use of $\ell_2$-norm regularization.
The result below holds with
high probability with respect to a random permutation of data. 
\bet\label{thm:complexity}
For any $n_0,\delta,\epsilon$ there exists $\epsilon_0$ s.t.
data-access complexity of Algorithm \ref{alg:optimal} is
\[
  \Ocal\left(\frac{\kappa}{\lambda\epsilon}\cdot
   (\log\log(1/\epsilon) + \log(1/\delta))\right)\] 
and $\fh(\wb_T)-\fh(\wbh^*)\leq \epsilon$ w.p. at least $1-\delta$.
\eet

\proof 
Recall that we denote the approximation error estimate at stage $t$ by
$\gh_t(\wb)\triangleq\fh_t(\wb)-\fh_t(\wbh_t^*)$ and the full
approximation error as $\gh(\wb)\triangleq \fh(\wb)-\fh(\wbh^*)$.
Moreover, note that Algorithm \ref{alg:optimal} defines
the optimization error tolerances recursively as
$\epsilon_{t+1}=\epsilon_t/2$.
First, we give the following uniform convergence result:
\bel\label{lem:uniform-convergence}
For any $n_0,\delta, T$, there exists $\epsilon_0$ such that
$$\epsilon_0=\Ocal(L^2B^2\log(T/\delta)\cdot \lambda^{-1}),$$
 and with probability $1-\delta$, for all $\wb$ and all $0\leq t < T$:
\begin{align}
\gh_0(\wb_0) &\leq\epsilon_0,\label{eq:base-case}\\
\gh_{t+1}(\wb) &\leq 2 \cdot \gh_t(\wb) + \epsilon_{t},\label{eq:inductive-step}\\
\gh(\wb) & \leq 2\cdot \gh_T(\wb) + \epsilon_T.\label{eq:full-loss}
\end{align}
\eel
See Appendix \ref{sec:lemma-proof} 
\ifcitesupp~ of \cite{supplement} \fi
for proof.
Next, using this lemma, we show the main result.
Let $\kappa$ be the convergence rate of
the inner optimizer.
Recall, that we set the number of inner iterations
of Algorithm \ref{alg:optimal} to be
\begin{align*}
  \hat{\kappa} \triangleq \lceil \kappa \log(6) \rceil.
\end{align*}
This gives us the following bound for the progress that the inner
optimizer makes at each stage (for any $0\leq t\leq T$):
\begin{align*}
  \gh_{t+1}(\wb_{t+1})&\leq \left(1-\frac{1}{\kappa}\right)^{\hat{\kappa}}
  \gh_{t+1}(\wb_t)\\
  &\leq \exp\left(-\frac{\hat{\kappa}}{\kappa}\right)\,\gh_{t+1}(\wb_t)
  \leq \frac{\gh_{t+1}(\wb_t)}{6}.
\end{align*}
Suppose that $\epsilon_0$ satisfies Lemma
\ref{lem:uniform-convergence}. 
We can show by induction that (with
probability $1-\delta$) for all 
$t\leq T$, model $\wb_t$ is an 
$\epsilon_t$-approximate solution for $\fh_t$, i.e. that
$\gh_t(\wb_t)\leq\epsilon_t$. Base case is given by
(\ref{eq:base-case}). The inductive step follows from: 
\begin{align*}
\gh_{t+1}(\wb_{t+1}) &\leq 
\frac{2\cdot \gh_t(\wb_t) + \epsilon_t}{6} \leq \epsilon_{t+1}.
\end{align*}
Next, we verify that $\wb_T$ is an $\epsilon$-approximate solution for
$\fh$:
\[\gh(\wb_T)\leq 2\cdot \gh_T(\wb_T) +
\epsilon_T\leq 3\cdot \epsilon_T\leq\epsilon.\]
Finally, we move on to complexity analysis. The number of
iterations in the algorithm is $T=\Ocal(\log(\epsilon_0/\epsilon))$, since:
\begin{align*}
2^T &= \frac{2\cdot\epsilon_0}{\epsilon_{T-1}}<6\cdot \frac{\epsilon_0}{\epsilon}.
\end{align*}
Assuming that one update of the inner optimizer requires $C$ passes over the data, we obtain the data-access complexity:
\begin{align*}
\sum_{t=1}^T \hat{\kappa}\,C\,n_t &= C\,\hat{\kappa}\, n_0\sum_{t=1}^T 2^t
\\
&\leq 2C\,\hat{\kappa}\, n_0\cdot 2^T  = \Ocal\left(n_0\,\epsilon_0 \cdot
\frac{\kappa}{\epsilon}\right). 
\end{align*}
See Appendix \ref{sec:log-terms} 
\ifcitesupp~ of \cite{supplement} \fi 
for details regarding the log terms. 
\qed


Using gradient descent as the inner optimizer, we have $\kappa =
\Ocal(1/\lambda)$, so Algorithm \ref{alg:optimal} reaches data-access
complexity $\Ocalt(1/(\lambda^2\epsilon))$. However, the general nature of
this approach allows us to choose a different linear optimizer with
better guarantees, like $\kappa=\Ocal(1/\sqrt{\lambda})$ for accelerated
gradient descent. Methods like L-BFGS and other approximate Newton algorithms have
been shown to exhibit linear convergence
\cite{trust-region-large-scale,sub-sampled-newton-convergence} with 
a rate that does not suffer from such strict 
dependence on the strong convexity coefficient $\lambda$. Hence, when
using those optimizers, for most problems we should expect $\kappa$ to
be a small constant factor, 
in which
case data-access complexity becomes $\Ocalt(1/(\lambda\epsilon))$.

%% file: experiments.tex

\begin{figure}
\vspace{-5mm}
\includegraphics[width=0.5\textwidth]{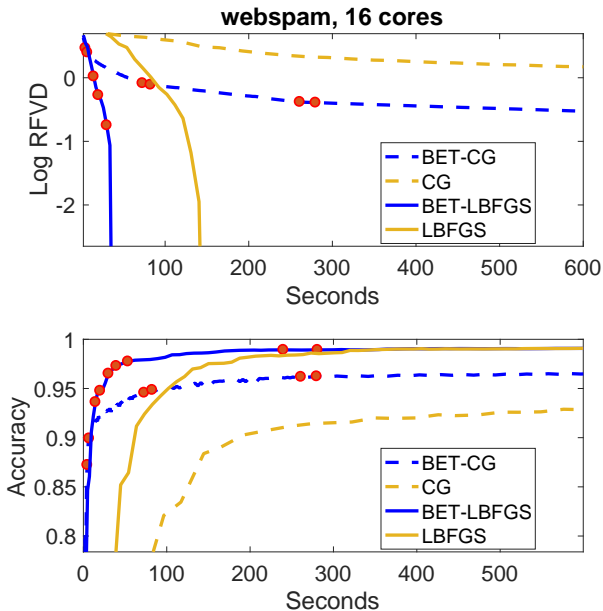}
\ifisarxiv
\vspace{-8mm}
\else
\vspace{-5mm}
\fi
\caption{Comparing BET and Batch, for two inner optimizers:
  Nonlinear-CG (CG) and Limited-memory BFGS (L-BFGS) (webspam dataset,
  16 cores). Circular dots in BET denote the batch expansion points.}
\label{fig:cg-comparison}
\end{figure}

In this section we present experimental results showing the
benefits of applying BET acceleration to batch optimizers and
demonstrating scalability of the method in parallel and
distributed settings. 
As the optimization problem we use logistic loss
with $\ell_2$-norm regularization trained on several standard
LIBSVM datasets (see Table \ref{tab:datasets}). 
All algorithms start with the initial 
model vector $\wb$ set to all zeros. BET
was implemented as shown in Algorithm \ref{alg:two-track}, without
any parameters that required tuning. The results are presented using test
set accuracy and training objective, the latter being shown in terms
of the log Relative Functional Value Difference
\vspace{-1mm}
\begin{align}
\textnormal{log RFVD:}\quad
\log\left((\fh(\wb)-\fh(\wbh^*))/\fh(\wbh^*)\right).\label{eq:log-rfvd}
\end{align}
All algorithms were implemented in the PETSc
\cite{petsc-web-page,petsc-user-ref,petsc-efficient} framework, with
the data split between multiple computing cores to achieve
parallelization speed-up (we used 16 cores in most experiments). For
splice-site  dataset\footnote{Due to resource
  constraints, we used a 1.5TB portion of the full 3TB splice-site
  dataset.}, data was additionally divided between multiple machines. 

\begin{table}[H]
\begin{center}
\begin{tabular}{c|c|c|c}
Dataset, size & Train/Test & Dim. & $\lambda$ \\
\hline
url, 1GB& 1.8M/0.5M & 3.2M & 1e-8\\
covtype, 19GB& 0.5M/69k & 170k & 1e-6 \\
webspam, 30GB& 250k/100k & 16.6M & 1e-6 \\
splice-site, 1.5TB& 25M/4.6M & 11.7M & 2e-10
\end{tabular}
\caption{A list of datasets and regularization used for the
  experiments. For covtype, features were expanded to all monomials of
  degree 3.}  
\label{tab:datasets}
\end{center}
\end{table}

\subsection{Adapting to inner optimizers}
\label{sec:inner-optimizers}


To demonstrate the flexibility of our framework, in this section we
use two different inner optimizers with BET: 
\ifisarxiv\vspace{-2mm}\fi
\begin{enumerate}
\item Nonlinear Conjugate Gradient method (CG), using Fletcher-Reeves \cite{fletcher-reeves}
  formula, with exact line-search;
\item Limited-memory BFGS (L-BFGS) \cite{lbfgs}.
\end{enumerate}
\ifisarxiv\vspace{-2mm}\fi
Both of those methods are linear optimizers that employ strategies for
enhancing the basic gradient descent direction. Note, that CG uses a
memory vector updated 
at each iteration. When the loss function changes from $\fh_t$ to
$\fh_{t+1}$, one might expect that memory vector to become invalid,
rendering CG ineffective. However, BET still proves very
effective at accelerating memory-based algorithms.

Figure \ref{fig:cg-comparison} shows the performance comparison of
using BET with the two inner optimizers on webspam dataset, contrasted
with both of them ran in regular batch mode. First, we can see that L-BFGS
is a much more effective optimizer than Nonlinear CG, thus in all
of the following experiments we used L-BFGS as the inner
optimizer. However, both methods significantly benefit from BET. In
fact, in the early phase of
the optimization, performance of BET is similar with either of the
optimizers. However, once batch expansion reaches close to the full
dataset, quality of the underlying optimizer starts to play an
important role. Circular dots on the BET plots in Figure
\ref{fig:cg-comparison} mark the points when batch size is doubled
during the optimization. We observed that the average number of
iterations per stage of BET is larger when the inner optimizer is
CG. This matches our expectation, since theory suggests that the
number of iterations per stage should be inversely proportional to the
convergence rate of the inner optimizer. Moreover, within one
optimization run, we saw significant fluctuations in the number of
iterations needed before each batch-expansion stage,
which means that there is no universal number of iterations per
stage which will work well in all settings. Thus, the
adaptive capabilities of the two-track algorithm are crucial to
achieving good performance of batch-expansion.

\subsection{Combining the benefits of batch and stochastic}

\begin{figure}
\vspace{-5mm}
\includegraphics[width=0.5\textwidth]{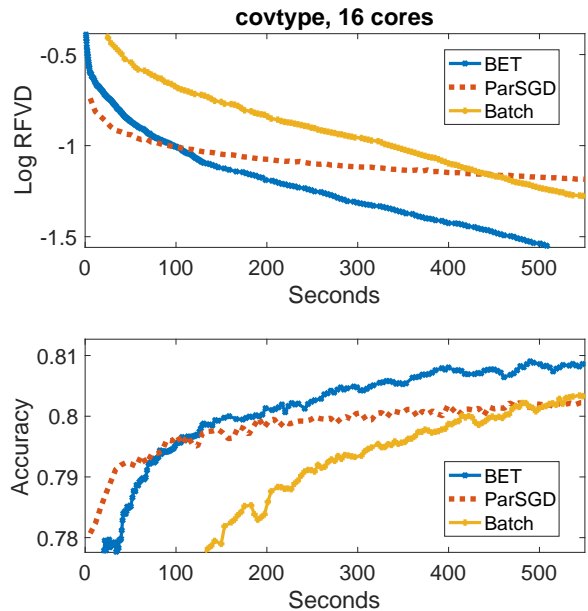}
\ifisarxiv
\vspace{-8mm}
\else
\vspace{-5mm}
\fi
\caption{Comparing BET, Batch (L-BFGS) and Parallel SGD for covtype dataset,
  running on 16 cores. More datasets are presented in Appendix
\ref{sec:batch-vs-sgd}\ifcitesupp~ of \cite{supplement}\fi.
} 
 \label{fig:covtype-tripled}
\end{figure}

In this experiment, we examine the trade-offs between batch and
stochastic methods, showing how BET fits into this
comparison. A batch optimizer typically starts off slower than a
stochastic one, but once it gets 
closer to the optimum, we expect to see a fast convergence behavior.
Thus, depending on which tolerance level we select, we would choose
the appropriate optimizer. This is shown on a sample plot for the
covtype dataset in Figure \ref{fig:covtype-tripled}, where we use Parallel
SGD \cite{parallel-sgd} as the stochastic algorithm and L-BFGS as the
Batch method, with all algorithms running on a 16 core machine (for
the sake of clarity, only a portion of the full convergence time is
shown). We also tested mini-batch SVRG as an alternative stochastic
approach. However, due to the high communication cost, this method
exhibited much slower per iteration time, and for this reason we did
not include the results in the paper. To obtain the reported
performance for Parallel SGD  we had to tune the
step size on a chunk of data, which adds to the overall optimization
time, whereas BET is a parameter-free method. Our algorithm gets the best of
both worlds, since it behaves
like a stochastic method at the beginning, and then like a batch
method towards the end. More plots are presented in Appendix
\ref{sec:batch-vs-sgd}\ifcitesupp~ of \cite{supplement}\fi,
showing datasets which favor either batch or stochastic methods.
In summary, BET performs as well as the best method for each dataset,
with no tuning necessary.



\subsection{Parallelization speed-up}

\begin{figure}
\vspace{-1.5mm}
\includegraphics[width=0.5\textwidth]{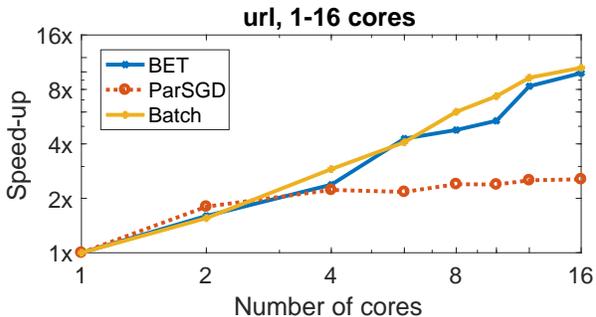}
\ifisarxiv
\vspace{-7mm}
\else
\vspace{-5mm}
\fi
\caption{Parallelization speed-up on url dataset for Batch, BET and Parallel
    SGD.}
\label{fig:speed-up}
\end{figure}

The goal of this experiment is to analyze the parallelization speed-up of BET
and compare it with the speed-up of Batch. We set this up as
follows: we run to convergence and find the final accuracy, then for
each method, we vary the number of cores and measure the time it takes
to reach within 0.25\% of optimum accuracy. For this experiment we
selected the url dataset to demonstrate a stark contrast with Parallel
SGD. The method achieves close to linear speed-up only for up to 4
cores, after which its performance flattens.
This behavior can be atributed to the fact that as we increase the
number of cores, a single iteration of Parallel
SGD becomes less and less effective, which in some cases may negate
the parallelization speed-up altogether. Batch methods, on the other
hand, behave much more reliably with parallelization, because their
iterations produce the same effect regardless of the number of cores.
Figure \ref{fig:speed-up} shows the speed-up factors for BET,
Batch and Parallel SGD. BET achieves similar speed-up as Batch,
due to the fact that parallelization happens in the inner
optimizer, which is the same for both methods.



\subsection{Distributed optimization}

\begin{figure}
\vspace{-3mm}
\includegraphics[width=0.5\textwidth]{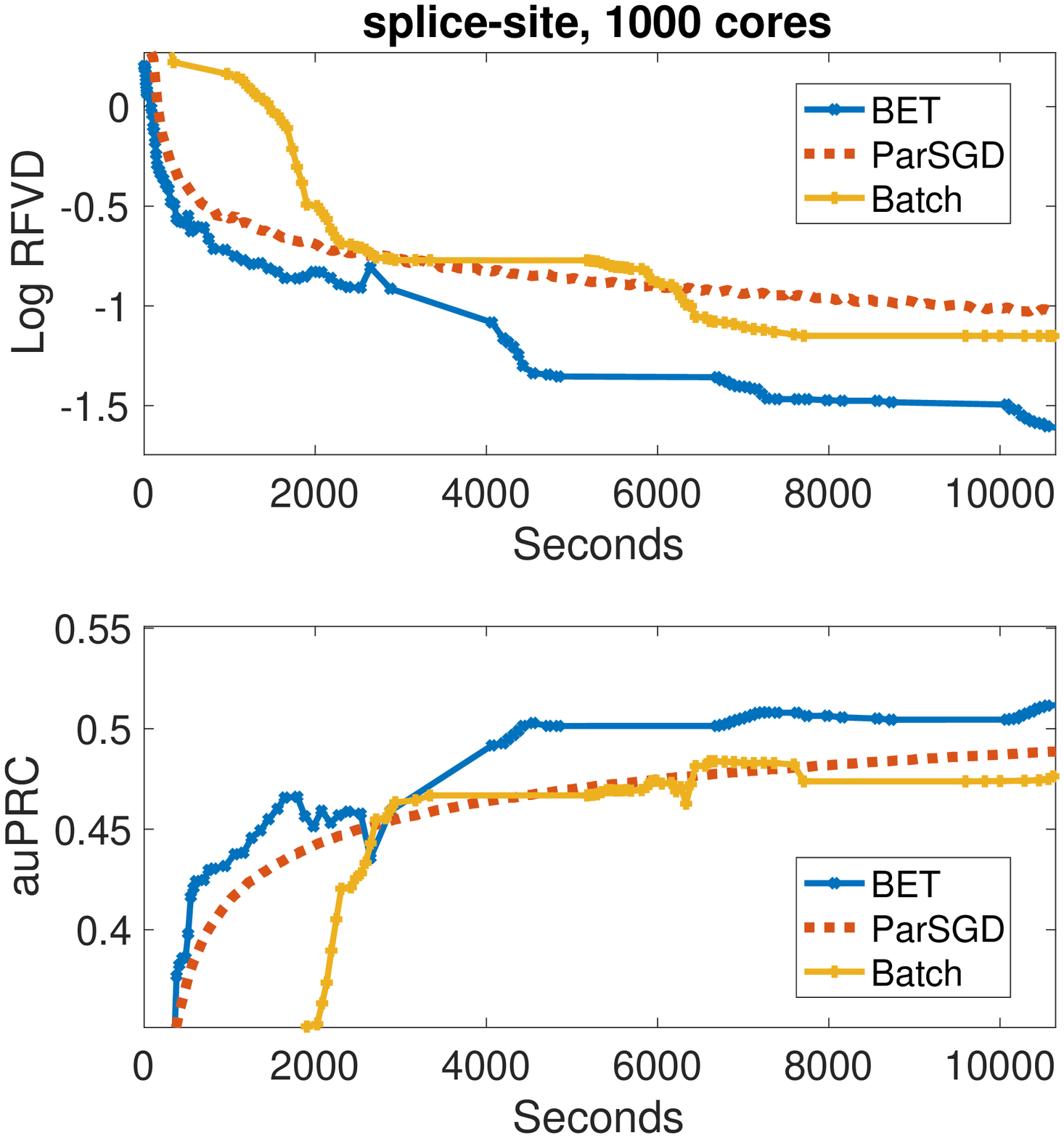}
\ifisarxiv
\vspace{-8mm}
\else
\vspace{-6mm}
\fi
\caption{Distributed experiment comparing BET to Batch (L-BFGS) and Parallel
  SGD on splice-site dataset.}
\label{fig:splice-site}
\end{figure}

Batch optimization shows the most benefits when dealing with
large-scale datasets, which do not fit into the memory of a single
machine. BET easily scales up to this setting as seen
in Figure \ref{fig:splice-site}, running on the splice-site data,
with 20 machines, and 50 cores per machine, compared against Batch and
Parallel SGD in the same setup. Since the dataset is highly skewed,
the results are shown in terms of area under Precision-Recall curve
(auPRC).  

Note that BET reaches close to optimum test auPRC much more quickly
than Batch or Parallel SGD. Moreover, until that point batch-expansion
has not reached the full
training data, which means that if the data were loaded in parallel
with the optimization, the optimization time would be partially
absorbed into the loading time. In our experiments, each one of the
1000 processes independently reads a separate chunk of data, however
this procedure still takes at 
least 1500 seconds even on a high-performance cluster. Thus, a portion
of BET's optimization time can be absorbed into data loading.

%% file: conclusions.tex
We proposed Batch-Expansion Training, a simple parameter-free method
of accelerating batch optimizers in a way that is both theoretically and
experimentally efficient. BET does not require tuning and can be
paired with any optimizer, offering advantages in parallel and
distributed settings. 
Extending our framework to non-convex optimization is an interesting
direction for future research.

%% file: lemma-proof.tex
\subsection{Proof of Lemma \ref{lem:uniform-convergence}}
\label{sec:lemma-proof}
First, we formulate a uniform-convergence bound, which closely
resembles Theorem 1 from \cite{fast-rates}. The only difference is
that they consider PAC setting: sampling an
i.i.d. dataset $\Zb$ from a fixed distribution,  and comparing the
finite-sample objective $\fh$ computed using $\Zb$ with the true
objective $f$, which is an expectation over the distribution. On the
other hand, we consider an increasing sequence of datasets
$\Zb_t=\{z_i\}_{i=1}^{n_t}$, selected by a uniformly random
permutation of the full dataset $\Zb$. Note, that we assume the 
algorithm never observes the full dataset, only loading as much data
as needed. Taking the limit of $N\rightarrow \infty$, the relationship
between any subset $\Zb_t$ and the full dataset $Z$
becomes statistically equivalent to i.i.d. sampling from any fixed underlying
distribution. Given that our goal is generalization to predicting on
new data, that simplification is reasonable, although the analysis does
go through in the strict optimization setting, where $N$ is finite.
However, even with this assumption, we still need to 
describe the relationship between two consecutive subsets in the
sequence, which does not fit the i.i.d. sampling model. To that
end, we can 
view $\Zb_t$ as a fraction of elements from $\Zb_{t+1}$, selected
uniformly at random without replacement. We now describe the
relationship between the two consecutive loss estimates in this
sequence. Note, that in this section the big-$\Ocal$ notation hides
only fixed numeric constants.

\bel\label{lem:fast-rates}
With probability $1-\delta$, for all $\wb$ and all $0\leq t\leq T$ we have
\begin{align}
\gh_{t+1}(\wb)\leq 2\, \gh_t(\wb) + \Ocal\left( \frac{ L^2B^2
    \log(T/\delta) }{\lambda\, n_{t}} \right). \label{eq:fast-rates}
\end{align}
\eel
\proof
The proof is very similar to \cite{fast-rates}, except we
replace standard Rademacher Complexity with Permutational
Rademacher Complexity (PRC), proposed in
\cite{permutational-rademacher}.
Let us fix $t$, and consider a specific set of instances $\Zb_{t+1}$,
from which a random subset $\Zb_t$ is sampled (without replacement).
Following \cite{fast-rates}, for any
$r>0$ we define 
\begin{align*}
\Hcal_{t,r} \triangleq \left\{h_{\wb}^{t,r}=\frac{h_{\wb}^t}{4^{k_{t,r}(\wb)}}\,:\,\wb\in\Wb \right\},
\end{align*}
where
\[k_{t,r}(\wb) \triangleq \min\{k'\in\zz_+\,:\,\fh_{t+1}(\wb)\leq r4^{k'}\}\]
and
\[h_{\wb}^t(z) \triangleq \ell_z(\wb) - \ell_z(\wbh_{t+1}^*).\]
Our aim is to analyze the empirical average of the function values
from $\cH_{t,r}$ evaluated on a given instance set $\Zb$:
$$\bar{h}^{t,r}_{\Zb}(\wb)=\frac{1}{|\Zb|}\sum_{z\in\Zb}h_{\wb}^{t,r}(z).$$
We can translate the task of comparing $\gh_{t+1}$ and $\gh_t$ to describe it in terms of the function class $\cH_{t,r}$:
\begin{align*}
  \gh_{t+1}(\wb) - \gh_t(\wb) &= \gh_{t+1}(\wb) - (\fh_t(\wb) -
  \fh_t(\wbh_t^*))\\
  &\leq \gh_{t+1}(\wb) - (\fh_t(\wb) - \fh_t(\wbh_{t+1}^*))\\
  &=4^{k_r(\wb)}\left[\bar{h}_{\Zb_{t+1}}^{t,r}(\wb) - \bar{h}_{\Zb_t}^{t,r}(\wb)\right].
\end{align*}
To compare $\bar{h}^{t,r}_{\Zb_t}$ with $\bar{h}^{t,r}_{\Zb_{t+1}}$
we use Theorem 5 \cite{permutational-rademacher}, which provides
transductive risk bounds through expected PRC of function class
$\Hcal_{t,r}$, conditioned on set $\Zb_{t+1}$: 
\[\Qcal(\Hcal_{t,r},\Zb_{t+1}) \triangleq
\ee\left[\hat{Q}_{n_t,n_t/2}(\Hcal_{t,r},\Zb_t)\,|\,\Zb_{t+1}\right].\] 
Here, the randomness only comes from selecting $\Zb_t$ as a subset
of $\Zb_{t+1}$. For any $\delta>0$, with probability at least $1-\delta$,
\begin{align*}
\sup_{\wb}&\left[\bar{h}_{\Zb_{t+1}}^{t,r}(\wb) - \bar{h}_{\Zb_t}^{t,r}(\wb)\right]\\
&\leq \underbrace{\Qcal(\Hcal_{t,r},\Zb_{t+1})}_{Y_1} +\underbrace{\sup_{h_{\wb}^{t,r},z}|h_{\wb}^r(z)|\cdot
\Ocal\left(\sqrt{\frac{\log(1/\delta)}{n_{t}}}\right)}_{Y_2}.
\end{align*}
Note, that $\Qcal(\Hcal_{t,r},\Zb_{t+1})=\Ocal(\Rcal_{n_t}(\Hcal_{t,r}))$ (see
\cite{permutational-rademacher}), where $\Rcal_{n_t}$ is the standard Rademacher Complexity.
The remainder of the proof proceeds
identically as in \cite{fast-rates} (up to numerical constants),
i.e. by bounding both terms $Y_1$ and $Y_2$ by
\[\Ocal\left(LB\sqrt{\frac{r\log(1/\delta)}{\lambda n_t}}\right).\]
Note, that since the bound is obtained for
every possible $\Zb_{t+1}$, it will still hold with probability at
least $1-\delta$ without conditioning on $\Zb_{t+1}$. 

Finally, as shown in \cite{fast-rates}, by setting $r$ appropriately
we obtain that w.p. $1-\delta$, for all $\wb$
\begin{align*}
\gh_{t+1}(\wb)\leq 2\, \gh_t(\wb) + \Ocal\left( \frac{ L^2B^2
    \log(1/\delta) }{\lambda\, n_{t}} \right).
\end{align*}
Applying union bound to account
for all values of $t$ simultaneously, we obtain the desired result.  
\qed

\begin{figure*}
\includegraphics[width=0.5\textwidth]{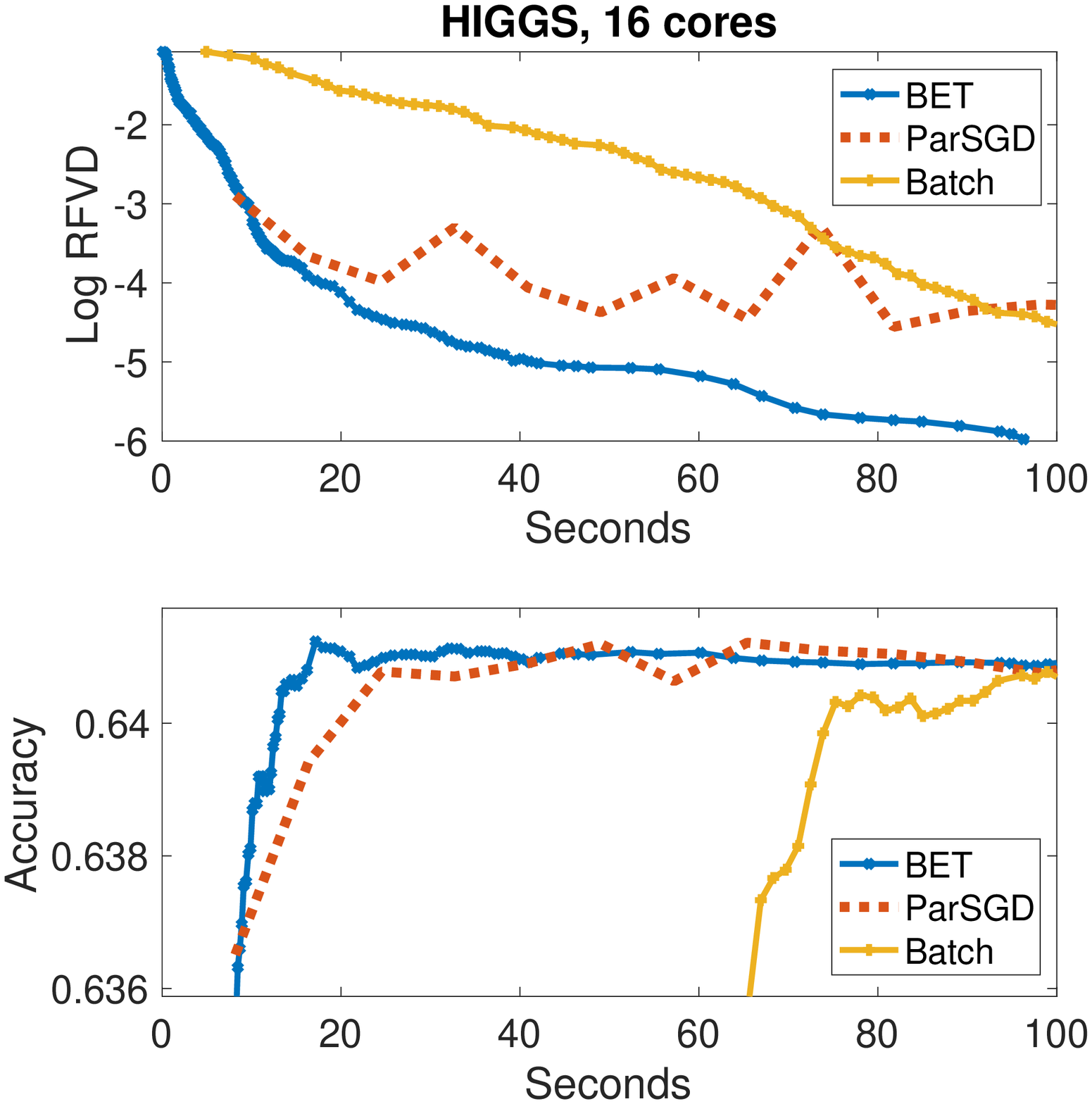}
\includegraphics[width=0.5\textwidth]{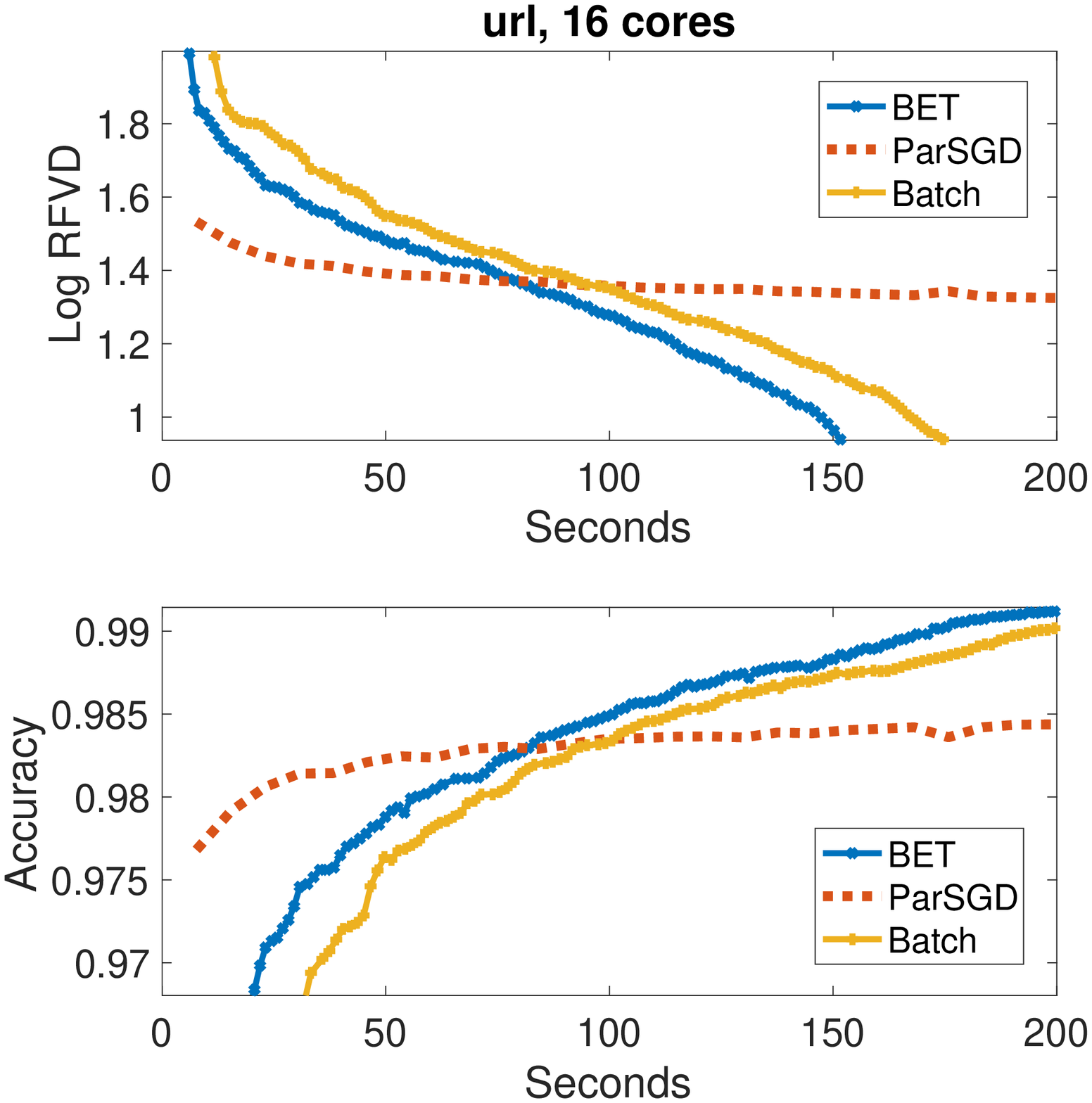}
\vspace{-5mm}
\caption{Comparing BET, Batch and Parallel SGD for HIGGS (left) and
  url (right) datasets,
  running on 16 cores. BET is as good as the best method in each case.} 
 \label{fig:additional}
\end{figure*}

We return to the proof of Lemma \ref{lem:uniform-convergence}.
Using Lipschitz and boundedness assumptions for the loss $\ell$ and
mapping $\phi$, as well as strong convexity of the regularized
objective, we obtain initial tolerance of the loss estimate: 
\begin{align*}
\gh_0(\wb_0)  &= \fh_0(\wb_0)-\fh_0(\wbh_0^*)\\
&\leq \frac{1}{n_0}\sum_{i=1}^{n_0}\left(\ell_{z_i}(\wb_0) -
\ell_{z_i}(\wbh_0^*)\right) \\
&\leq LB\|\wbh_0^*\!\| \leq LB \sqrt{\frac{2\,\gh_0(\wb_0)}{\lambda}},\\ 
\gh_0(\wb_0) &\leq\frac{2L^2B^2}{\lambda}.
\end{align*}
We used the fact that $\wb_0$ is set to zero only for applying inequality
$\|\wb_0\|\leq\|\wbh_0^*\!\|$ to drop the regularization terms
(any initialization satisfying that 
requirement is acceptable).

Finally, Condition (\ref{eq:full-loss}) regards the
relationship between approximation error estimate $\gh_T$ and full
approximation error $\gh$. This bound can be obtained by repeating the same
argument as in Lemma \ref{lem:fast-rates}. We can either assume
$N\rightarrow \infty$ and use standard Rademacher complexity, as in
Theorem 1, \cite{fast-rates}, or stay with the finite
optimization model and apply PRC. Thus, we can set
$\epsilon_0$ to satisfy the conditions of Lemma
\ref{lem:uniform-convergence}. \qed

\subsection{Deriving Log Terms in Theorem \ref{thm:complexity}}
\label{sec:log-terms}
The number of iterations, 
$T=\Ocal(\log(\epsilon_0/\epsilon))$, depends on $\epsilon_0$. But in
Lemma \ref{lem:uniform-convergence} we defined $\epsilon_0$ using $T$.
To address this, we have to find $\epsilon_0$ satisfying:
\begin{align*}
\epsilon_0\geq K\log\left(\frac{\log(\epsilon_0/\epsilon)}{\delta}\right),
\end{align*}
with $K=\Ocal(L^2B^2/\lambda)$. It is easy to show that for small
enough $\epsilon$ it suffices to set
\begin{align*}
\epsilon_0 &\triangleq
2K\log\left(\frac{\log(1/\epsilon)}{\delta}\right)\\
&=\Ocal \left(\frac{L^2B^2}{\lambda}\cdot (\log\log(1/\epsilon) +
  \log(1/\delta))\right). 
\end{align*}

Thus, setting $n_0=1$, we obtain the final complexity bound in Theorem \ref{thm:complexity} as
\[\Ocal \left(\frac{\kappa}{\lambda\epsilon}\cdot\,L^2B^2\cdot (\log\log(1/\epsilon) +
  \log(1/\delta))\right). \]

%% file: additional-experiments.tex
\label{sec:batch-vs-sgd}


\begin{table}[H]
\begin{center}
\begin{tabular}{c|c|c|c}
Dataset, size & Train/Test & Dim. & $\lambda$ \\
\hline
HIGGS, 8GB& 10.5M/0.5M & 28 & 1e-10 \\
url, 1GB& 1.8M/0.5M & 3.2M & 1e-8
\end{tabular}
\caption{A list of additional datasets and regularization used for the 
  experiments.}  
\label{tab:additional-datasets}
\end{center}
\end{table}

In this section we look at two datasets with very different
properties. First one, HIGGS, is large, but extremely low
dimensional. In this case, given an overabundance of data, if we look
at the accuracy plot (see Figure \ref{fig:additional}),
the Batch algorithm takes much longer to converge than Parallel SGD. This follows
from the fact that the task has low sample complexity, and a Batch
method is wasting resources by training on too much data.
The second dataset, url, is very high-dimensional, and in this case
Batch has clear advantage over SGD. BET does as well as the best
method on each 
dataset. In the case of HIGGS, BET simply converges to the optimum
accuracy before it even reaches full dataset, thus saving on expensive
iterations. For url, the batch expansion happens relatively early on
in the optimization, and from that point on the algorithm is simply
running full L-BFGS. Those two extreme cases show the versatility and
robustness of our proposed meta-algorithm.